\documentclass[conference]{IEEEtran}
\IEEEoverridecommandlockouts
\usepackage{cite}
\usepackage{amsmath,amssymb,amsfonts}
\usepackage{algorithmic}
\usepackage{graphicx}
\usepackage{textcomp}
\usepackage{xcolor}

\usepackage{booktabs}
\usepackage{makecell}

\def\BibTeX{{\rm B\kern-.05em{\sc i\kern-.025em b}\kern-.08em
 T\kern-.1667em\lower.7ex\hbox{E}\kern-.125emX}}

\begin{document}

\title{Roller-Quadrotor: \\A Novel Hybrid Terrestrial/Aerial Quadrotor \\with Unicycle-Driven and Rotor-Assisted Turning}
\author{Zhi Zheng$^{1,2,3}$, Jin Wang$^\dag$$^{1,2,3}$, Yuze Wu$^{4,5}$, Qifeng Cai$^{1,2,3}$, Huan Yu$^{1,2,3}$, \\Ruibin Zhang$^{4,5}$, Jie Tu$^{1,2,3}$, Jun Meng$^{2,3,6}$, Guodong Lu$^{1,2,3}$, and Fei Gao$^{4,5}$
	\thanks{This work was supported in part by the National Natural Science Foundation of China under Grant 52175032, the ``Pioneer" and ``Leading Goose" R\&D Program of Zhejiang under Grant 2023C01070, and Robotics Institute of Zhejiang University under Grant K12107 and K11805.}
	\thanks{
	$^{1}$The State Key Laboratory of Fluid Power and Mechatronic Systems, School of Mechanical Engineering, Zhejiang University, Hangzhou 310027, China. $^{2}$Robotics Institute of Zhejiang University, Hangzhou 310027, China. $^{3}$Robotics Research Center of Yuyao City, Ningbo 315400, China. $^{4}$The State Key Laboratory of Industrial Control Technology, Zhejiang University, Hangzhou 310027, China. $^{5}$Huzhou Institute of Zhejiang University, Huzhou 313000, China. $^{6}$Center for Data Mining and Systems Biology, College of Electrical Engineering, Zhejiang University, Hangzhou 310027, China.
	}
	\thanks{Email: {\tt\small\{z.z, dwjcom\}@zju.edu.cn}}
	\thanks{\dag Corresponding author: Jin Wang.}
}
\maketitle

\begin{abstract}
The Roller-Quadrotor is a novel quadrotor that combines the maneuverability of aerial drones with the endurance of ground vehicles. This work focuses on the design, modeling, and experimental validation of the Roller-Quadrotor. Flight capabilities are achieved through a quadrotor configuration, with four thrust-providing actuators. Additionally, rolling motion is facilitated by a unicycle-driven and rotor-assisted turning structure. By utilizing terrestrial locomotion, the vehicle can overcome rolling and turning resistance, thereby conserving energy compared to its flight mode. This innovative approach not only tackles the inherent challenges of traditional rotorcraft but also enables the vehicle to navigate through narrow gaps and overcome obstacles by taking advantage of its aerial mobility. We develop comprehensive models and controllers for the Roller-Quadrotor and validate their performance through experiments. The results demonstrate its seamless transition between aerial and terrestrial locomotion, as well as its ability to safely navigate through gaps half the size of its diameter. Moreover, the terrestrial range of the vehicle is approximately 2.8 times greater, while the operating time is about 41.2 times longer compared to its aerial capabilities. These findings underscore the feasibility and effectiveness of the proposed structure and control mechanisms for efficient navigation through challenging terrains while conserving energy.
\end{abstract}


\section{Introduction}
In recent years, unmanned aerial vehicles (UAVs) have witnessed growing utilization across diverse domains, including military operations, exploration missions, and search and rescue endeavors \cite{cite:b1}. However, these applications present formidable challenges to UAVs, particularly in terms of energy consumption \cite{cite:b2} and navigation through specialized terrains, especially narrow gaps \cite{cite:b3}. In order to overcome these obstacles, it becomes imperative to develop advanced technologies that enhance the endurance and adaptability of UAVs to various terrains. This technological advancement holds tremendous potential for expanding the scope of UAV applications. Regrettably, conventional UAV optimization falls short in simultaneously addressing these multifaceted challenges.

\begin{figure}[t]
	\begin{center}
		\includegraphics[width=1.0\columnwidth]{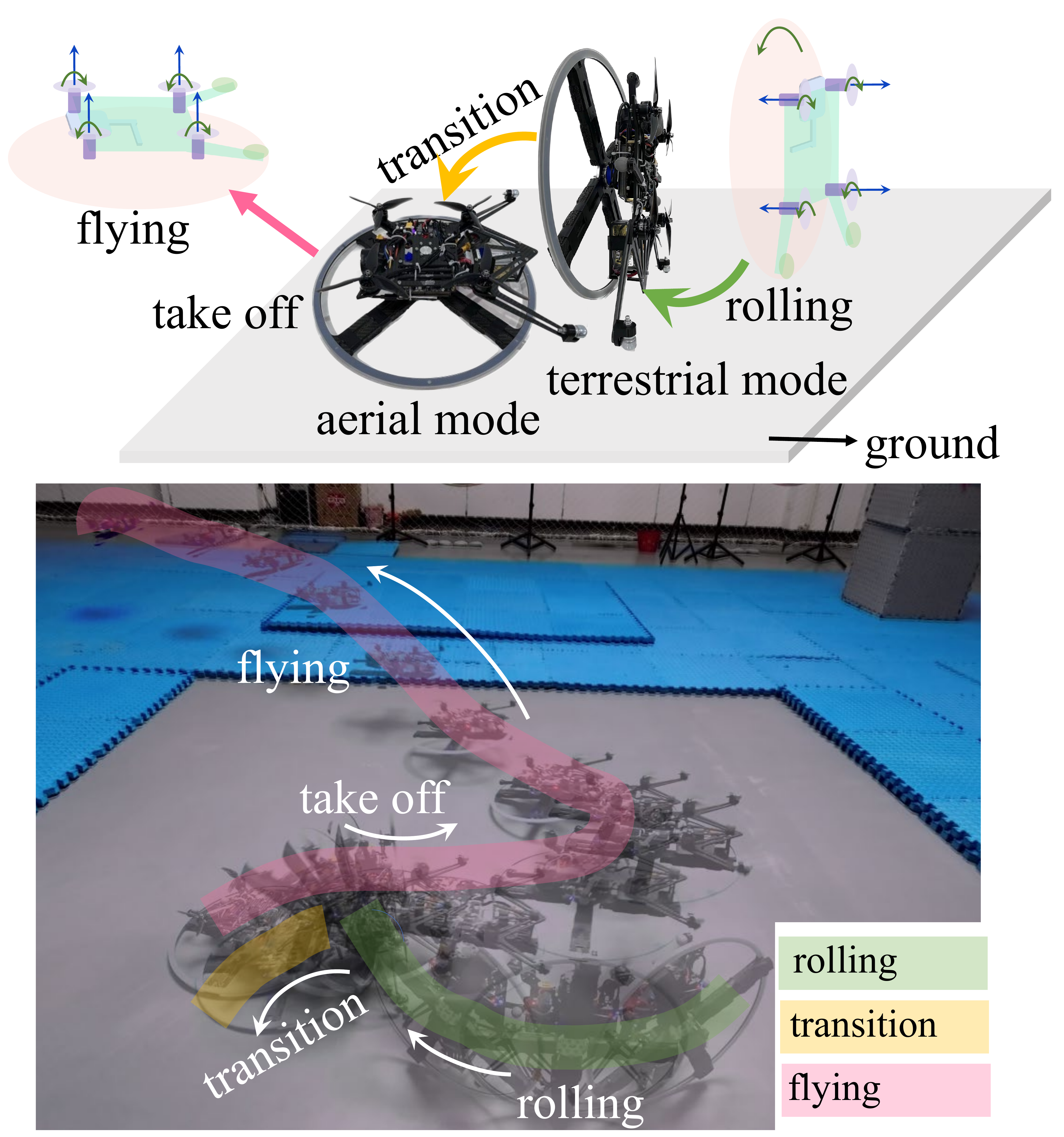}
	\end{center}
	\caption{
		\label{fig:fig1}
		We have devised multiple operational modes for the vehicle, encompassing rolling, transition, and flying. This diagram elucidates the practical implementation of these modes in real-world scenarios.
	}
\end{figure}

Zhang et al. have proposed an innovative solution to address energy consumption concerns, wherein quadrotor pitching is utilized to generate forward thrust \cite{cite:b4} \cite{cite:b5} \cite{cite:b6}. This approach allows for the control of passive wheels on the ground, enabling the vehicle to roll and overcome obstacles during flight. A similar concept has been explored in the works of Kalantari et al. and Dudley et al \cite{cite:b7} \cite{cite:b8} \cite{cite:b9}. However, it is crucial to consider the influence of ground effect (specifically, Wing-In-Ground effect or Wing-In-Surface-Effect) as the aircraft approaches the ground \cite{cite:b10}. The presence of turbulent airflow around the rotors in this scenario can result in changes in lift and drag, directly impacting the precision of motion control. Moreover, the ground effect can pose significant challenges to the design of control models and controllers. Additionally, the incorporation of wheels perpendicular to the UAV frame plane introduces notable increases in size and weight, subsequently compromising maneuverability. In summary, although the quadrotor and differential wheel approach can reduce energy consumption, it also yields adverse effects, rendering widespread implementation challenging.

Numerous endeavors have been undertaken to enhance the adaptability of vehicles to different terrains. In certain application scenarios such as inspection and exploration, vehicles may encounter the challenge of navigating through narrow gaps and passages, including square ventilation ducts and urban sewer pipes. Such scenarios pose significant obstacles to UAV applications. To address this issue, Falanga et al. have proposed foldable mechanical designs for the airframe \cite{cite:b11} \cite{cite:b12} \cite{cite:b13}. Their concept involves a morphing quadrotor, which allows the robot to fold its structure and reduce its overall span, facilitating passage through narrow apertures \cite{cite:b14}. Similarly, Bucki et al. have proposed a bird-inspired robot with passive joints that can temporarily reduce propulsion commands, thereby compressing its footprint to traverse narrow gaps \cite{cite:b15}. However, the adoption of folding mechanisms introduces a set of complex mechanical structures that give rise to several challenges. Firstly, the system exhibits high nonlinearity, complicating control strategies. Secondly, the intricate assembly of various components increases vibration and introduces uncertain noise. Lastly, the presence of numerous degrees of freedom results in cumulative errors, rendering control more challenging. In conclusion, while folding mechanisms prove beneficial for traversing narrow gaps, their implementation is hindered by various disadvantages, making practical application difficult.

\begin{figure}[t]
	\begin{center}
		\includegraphics[width=1.0\columnwidth]{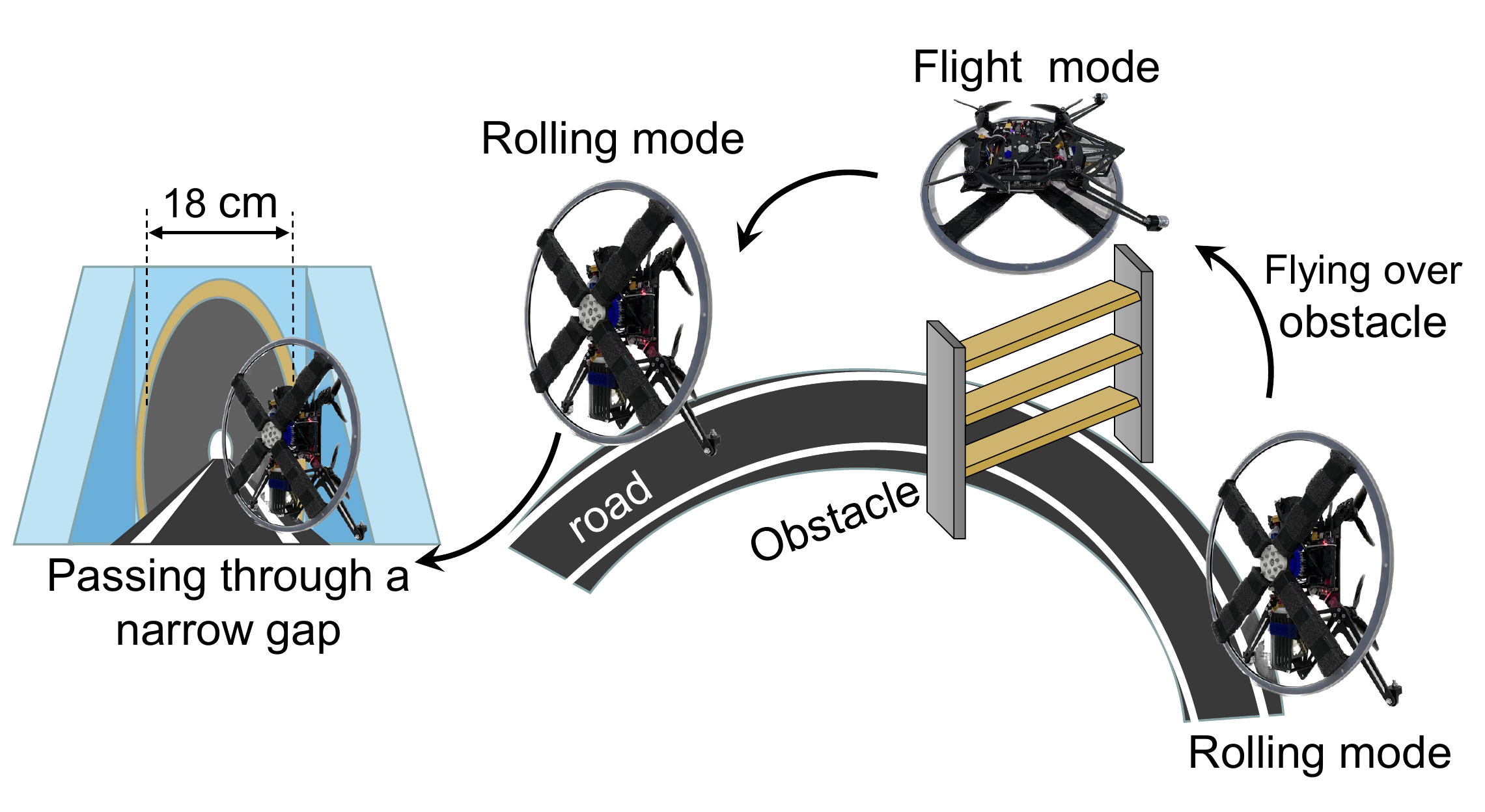}
	\end{center}
	\caption{
		\label{fig:fig2}
	The figure illustrates the schematic diagram showcasing the versatile capabilities of the Roller-Quadrotor, encompassing rolling on road surfaces, aerial flight over obstacles, and successfully passing through narrow gaps.
	}
\end{figure}

In summary, it is evident that a singular technology capable of simultaneously addressing multiple challenges remains elusive. Consequently, to tackle these issues comprehensively, we introduce a groundbreaking solution in the form of the Roller-Quadrotor, a nonfolding quadrotor equipped with a driving wheel. This novel hybrid vehicle combines terrestrial and aerial capabilities, facilitating versatile functionality. Notably, the Roller-Quadrotor adopts an unprecedented approach by utilizing a single wheel parallel to the frame for driving the rolling mode. Additionally, we have devised an innovative turning mechanism that harnesses rotor thrust. These advancements bestow upon the Roller-Quadrotor a distinctive advantage over conventional UAVs, namely, significantly reduced energy consumption.

We have developed comprehensive models and controllers to facilitate the rolling and transition modes of the vehicle during both aerial and terrestrial locomotion. These advancements are built upon the planar unicycle motion model and the first-order inverted pendulum model, representing notable improvements in terms of vehicle dynamics. To assess the functionality and performance of the proposed vehicle, extensive testing has been conducted in real-world environments. Comparative evaluations with other existing vehicles have been performed to highlight the innovation and feasibility of our design.

In terms of energy consumption, the Roller-Quadrotor demonstrates remarkable capabilities. Its terrestrial range surpasses the aerial range by approximately 2.8 times, while the operating time extends by an impressive factor of 41.2 compared to its aerial counterpart. Furthermore, when compared to another hybrid terrestrial and aerial quadrotor \cite{cite:b7} \cite{cite:b8}, the Roller-Quadrotor showcases superior performance by achieving a longer operating time in rolling mode, despite having the same vehicle mass. With regards to terrain adaptability, particularly in rolling through narrow gaps, the vehicle excels with an outstanding aspect ratio of approximately 3. In rolling mode, it possesses a diameter of 36 cm and a width of 12 cm. This exceptional design allows the Roller-Quadrotor to maneuver through narrow gaps half the size of its diameter, surpassing the capabilities of other similar vehicles.

The contribution of the proposed Roller-Quadrotor is summarized as follows:

\begin{itemize}
	\item Propose a novel hybrid terrestrial and aerial quadrotor featuring a novel unicycle-driven and rotor-assisted turning structure. This structure enables the vehicle to efficiently roll during terrestrial locomotion, conserving energy and enhancing terrain adaptability, particularly in navigating narrow gaps.
	\item Develop optimized models and controllers for vehicle rolling and transitioning between aerial and terrestrial locomotion, utilizing the motion model of the plane unicycle and the first-order inverted pendulum model.
	\item Conduct seven diverse experiments, demonstrating that the vehicle exhibits a terrestrial range approximately 2.8 times greater and an operating time about 41.2 times greater than its aerial range/operating time. Moreover, it successfully navigates through gaps half the size of its diameter, ensuring safe passage.
\end{itemize}

\section{Mechatronic system design}
This section first discusses the system architecture and components of Roller-Quadrotor mechatronic system, and then introduces the drive and transmission system design.

\begin{figure}[h]
	\begin{center}
		\includegraphics[width=1.0\columnwidth]{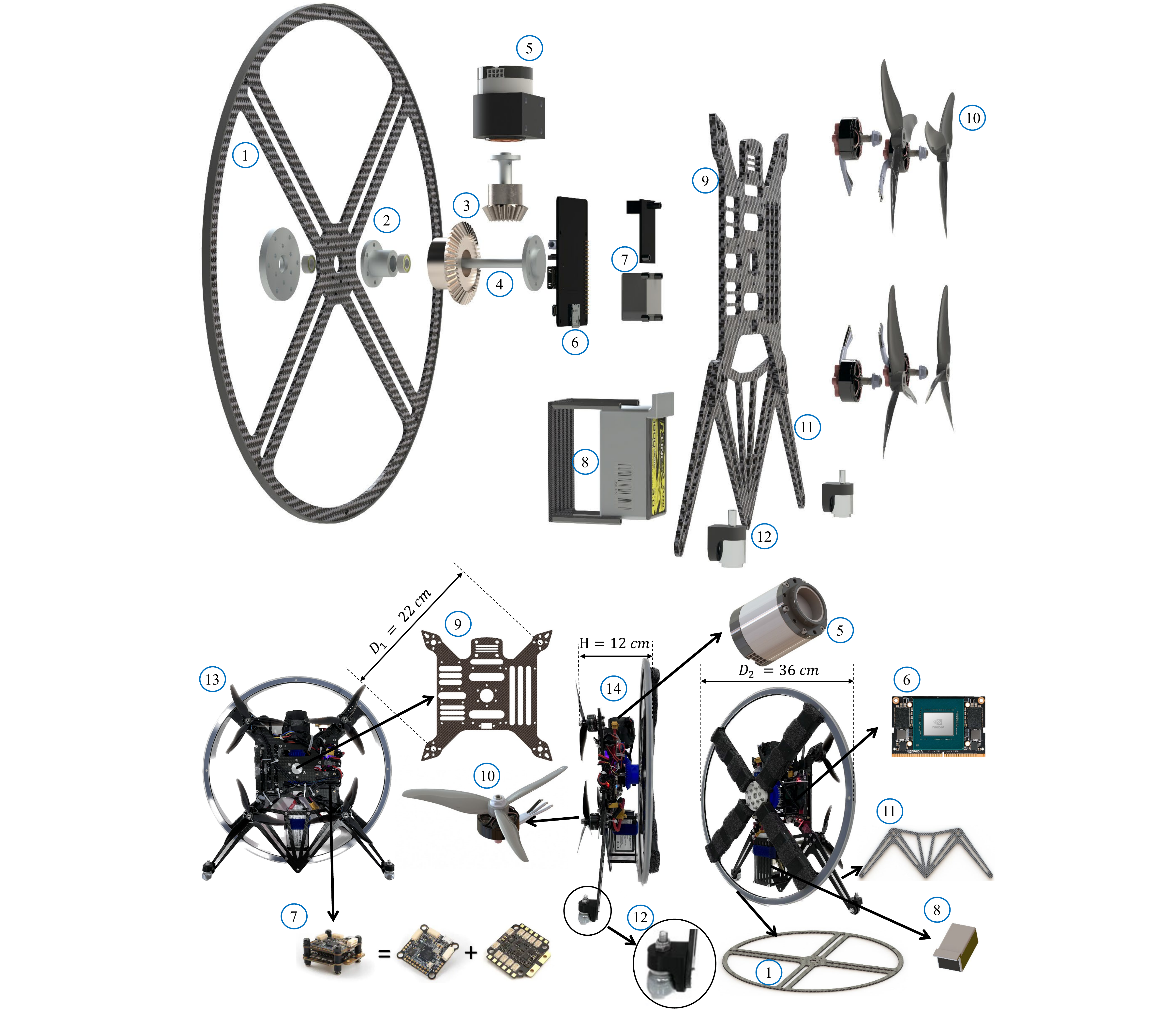}
	\end{center}
	\caption{
		\label{fig:fig3}
		The detailed composition of Roller-Quadrotor. The serial numbers represent (1) four-spoke wheel, (2) bearing, (3) bevel gears, (4) shaft, (5) servomotor, (6) onboard computer, (7) flight controller and electronic speed controller (ESC), (8) battery, (9) quadrotor frame, (10) rotors and 5-inch three-blade propellers, (11) frame support plate, (12) horn gimbals, (13) the top view of the actual vehicle, (14) lateral view of the actual vehicle.
	}
\end{figure}

\subsection{System Architecture and Components}
In the flight mode, Fig. \ref{fig:fig3} (13) presents a top view of the actual vehicle, offering insights into its physical configuration. The main body exhibits an asymmetrical design, featuring an X-shaped frame with a diameter of 22 $cm$ (refer to Fig. \ref{fig:fig3} (9)). To generate thrust, the vehicle utilizes T-MOTOR F60 PRO 2550KV rotors in conjunction with 5-inch three-blade propellers (depicted in Fig. \ref{fig:fig3} (10)). For flight control and electronic speed regulation, the Holybro Kakute H7 v1 flight controller and Tekko32 Metal 4 in 1 65A ESC STACK are employed (as depicted in Fig. \ref{fig:fig3} (7)).

To facilitate low-energy consumption in the rolling mode and enable passage through narrow gaps, a four-spoke wheel design is implemented as the primary component (refer to Fig. \ref{fig:fig3} (1)). This wheel possesses a diameter of 36 $cm$ and is equipped with a white rubber ring on its outer edge to enhance frictional force. Driving the wheel is a servomotor (depicted in Fig. \ref{fig:fig3} (5)) connected to the frame's shaft through a transmission system. The frame itself is linked to a frame support plate (illustrated in \ref{fig:fig3} (11)), where two horn gimbals (shown in Fig. \ref{fig:fig3} (12)) are mounted. These horn gimbals maintain contact with the ground, counterbalancing the reverse torque output of the motor during rolling and providing a mechanical constraint to ensure stability in the frame's pitch angle during the rolling process.

To power the system, a 2000 $mAh$ 4S battery (depicted in Fig. \ref{fig:fig3} (8)) is utilized. It is strategically positioned on the opposite side of the servomotor to ensure the stabilization of the body's center of gravity. As for the onboard computer, we have opted for the NVIDIA® Jetson Xavier NX (shown in Fig. \ref{fig:fig3} (6)). This high-performance computing platform is interconnected with the flight controller via the serial port to facilitate rotor thrust and reversing control. Additionally, it is connected to the servomotor through the CAN bus for torque and speed control. The decision to employ a potent graphics processing unit, such as the NVIDIA® Jetson Xavier, as the onboard computer was made in anticipation of future implementations involving deep learning and computer vision functionalities on the Roller-Quadrotor. This choice was prioritized over employing a more economical computer option like the Raspberry Pi.

To ensure optimal strength, the Roller-Quadrotor incorporates carbon fiber as the primary structural material. This material is utilized for key components such as the frame (refer to Fig. \ref{fig:fig3} (9)), the frame support plate (depicted in Fig. \ref{fig:fig3} (11)), and the four-spoke wheel (shown in Fig. \ref{fig:fig3} (1)). Additionally, a sponge strip is applied to the four-spoke wheel to provide cushioning and minimize impact during the transition to flight mode. This design choice enhances the overall robustness and durability of the system.

\subsection{Drive and Transmission System Design}
\begin{figure}[h]
	\begin{center}
		\includegraphics[width=1.0\columnwidth]{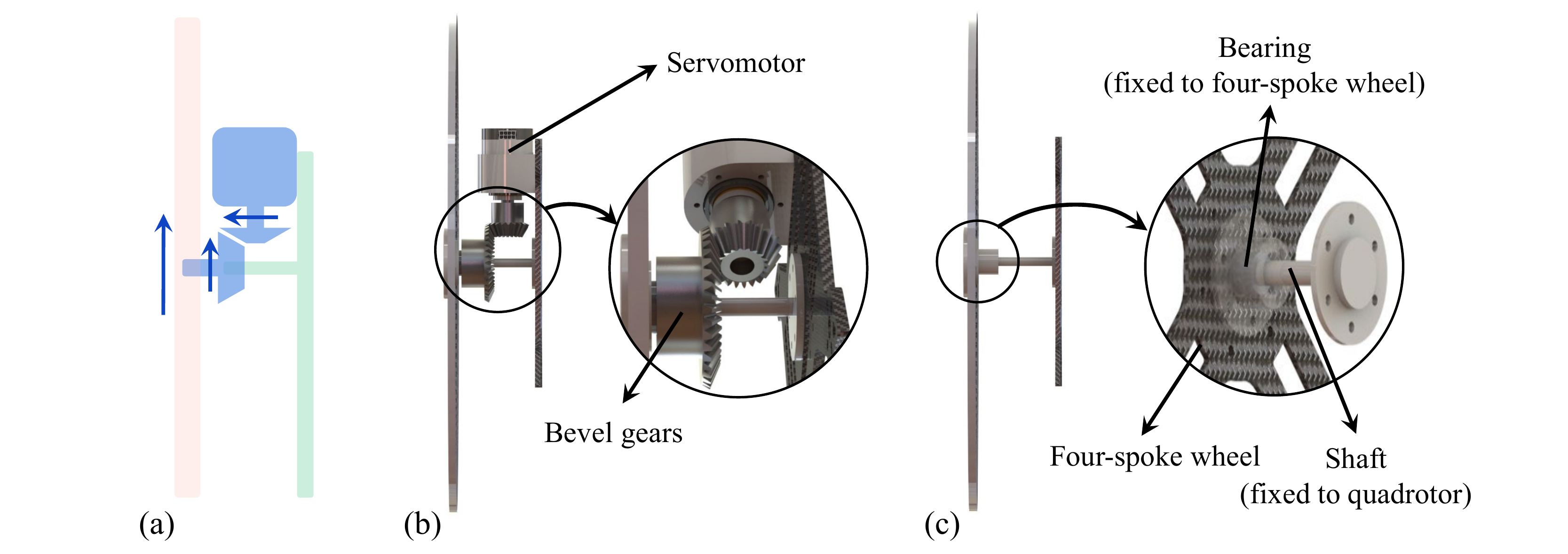}
	\end{center}
	\caption{
		\label{fig:fig4}
		Drive and transmission system design of rolling mode.
	}
\end{figure}

Upon activation of the rolling mode, the servomotor initiates the generation of torque, inducing rolling rotation around the initial yaw axis. This torque is subsequently transmitted through bevel gears (refer to Fig. \ref{fig:fig4}(b)) with a transmission ratio denoted as $i = 2$. The four-spoke wheel is securely affixed to the frame's fixed shaft and connected to the frame via bearings (depicted in Fig. \ref{fig:fig4}(c)). These components work in unison to facilitate smooth and controlled rolling motion during the terrestrial locomotion phase of the Roller-Quadrotor.

\vspace{-0.5cm}
\vspace{-0.2cm}

Simultaneously, the four rotors (refer to Fig. \ref{fig:fig8}) generate differential thrust, thereby imparting yaw torque to the system. This unique integration of ground rolling and differential thrust empowers the vehicle with enhanced versatility and efficiency during terrestrial locomotion, surpassing the capabilities of previous rolling vehicles. Notably, this approach substantially mitigates energy consumption, rendering it highly suitable for traversing narrow gaps with ease and agility. By synergistically leveraging these mechanisms, the proposed Roller-Quadrotor exhibits superior adaptability and energy efficiency in its ground motion capabilities.

\section{MODELING AND CONTROL}
In this section, we will comprehensively examine the different configurations of the Roller-Quadrotor and provide a sequential discussion on the corresponding modeling for each configuration.

\subsection{Rotor Thrus/Torque Model}
The thrust and torque generated by the $i$-th rotor (where $i\in {1, 2, 3, 4}$) will be denoted as $F_i$ and $\tau_i$, respectively. In order to ensure precise control of vehicles during non-flying modes, the relationship between the thrust $F_{i}$ and torque $\tau_i$ generated by each rotor is determined as a function of the rotational speed $\omega_{i}$.
\begin{equation}
	\label{eq:eq1}
	\begin{cases}
		F_{i} = \kappa_{f} * (\omega_{i})^2 \\
		\tau_{i} = \kappa_{m} * (\omega_{i})^2
	\end{cases}
\end{equation}
The equation (Eq. \ref{eq:eq1}) reveals that the thrust $F_{i}$ is contingent upon the coefficient of thrust, denoted as $\kappa_{f}$, while the torque $\tau_{i}$ is contingent upon the coefficient of torque, represented as $\kappa_{m}$.

During the non-flight mode of the vehicle, the thrust $F_{i}$ and torque $\tau_{i}$ generated by the rotors are not utilized for hovering or flying purposes. In such scenarios, we anticipate a relatively minor thrust $F_{i}$ from each rotor, and controlling the torque $\tau_{i}$ becomes unnecessary. Consequently, the angular speed $\omega_{i}$ demonstrates a linear relationship with the square of the thrust $F_{i}$ and torque $\tau_{i}$. It is reasonable to assume that the angular speed $\omega_{i}$ can be adequately maintained, allowing us to treat the coefficients of thrust $\kappa_{f}$ and torque $\kappa_{m}$ as constants.

The gravity of the vehicle and the angular speed $\omega_{i}$ of the rotors during hover are measured. The coefficient of thrust $\kappa_{f}$ is determined using Eq. \ref{eq:eq1}, while the coefficient of torque $\kappa_{m}$ is obtained through a torque measuring instrument.

The flight controller effectively regulates the angular speed $\omega_{i}$ by establishing communication with the Electronic Speed Controller (ESC) through the dshot protocol. To model the relationship between the angular speed $\omega_{i}$ and the dshot signal $U_d$, we employ a quadratic function, given by:
\begin{equation}
	\label{eq:eq2}
	\omega_{i} = {p}_{1} * U_d^2 + {p}_{2} * U_d + {p}_{3}
\end{equation}
The equation (Eq. \ref{eq:eq2}) reveals that the speed $\omega_{i}$ is influenced by several parameters, specifically denoted as ${p}_{1}$, ${p}_{2}$, and ${p}_{3}$.

By extracting the angular speed $\omega_{i}$ and dshot $U_d$ data from the flight controller's log, we can perform a fitting procedure using Equation \ref{eq:eq2} to determine the coefficients ${p}_{1}$, ${p}_{2}$, and ${p}_{3}$. Subsequently, we can utilize these coefficients to control the rotor angular velocity $\omega_{i}$ by providing the desired dshot values $U_d$ for each of the four rotors. This control mechanism allows us to obtain the ultimate outputs of thrust $F_{i}$ and torque $\kappa_{m}$.

\subsection{Wheel Torque/Force Model}

To ensure precise control of the torque $\tau_{w}$ and force $F_{w}$ applied to the wheel during non-flying modes, we can employ the mechanical drive structure to compute the wheel output using the following equation:
\begin{equation}
	\label{eq:eq3}
	\begin{cases}
		P_{s} = \tau_{s}*\omega_{s} \\
		\omega_{w} = \omega_{s} / i\\
		\nu_{w} = \omega_{w} * R_{w}\\
		\tau_{w} = \tau_{s} / i \\
		F_{w} = \tau_{w} / R_{w}\\
	\end{cases}
\end{equation}
In the equation, $P_{s}$ represents the output power of the servomotor, while $R_{w}$ and $i$ denote the constants representing the radius of the wheel and the transmission ratio, respectively.

From Eq. \ref{eq:eq3}, it becomes evident that by effectively regulating the angular velocity $\omega_{s}$ and torque $\tau_{s}$ output of the servomotor, we can precisely control the angular velocity $\omega_{w}$, linear velocity $\nu_{w}$, torque $\tau_{w}$, and force $F_{w}$ exerted by the wheel.

\subsection{Flight Model}
In this section, we will elaborate on the operational characteristics of the Roller-Quadrotor during take-off and flight, where four rotors are utilized for both lift and propulsion. This configuration enables the Roller-Quadrotor to navigate within a three-dimensional space, execute stable hovering maneuvers, and achieve rapid and agile movement in any desired direction.

\begin{figure}[h]
	\begin{center}
		\includegraphics[width=1.0\columnwidth]{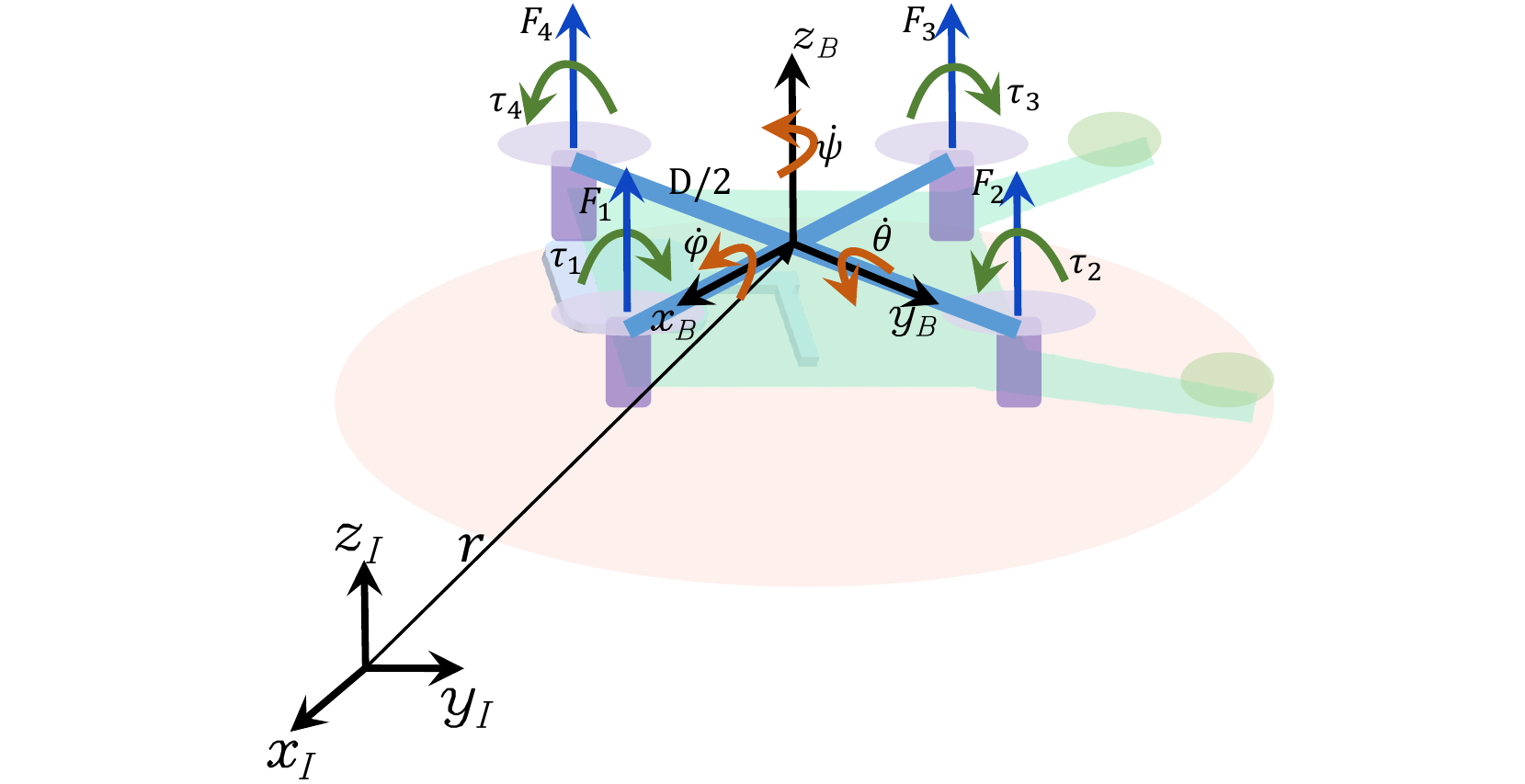}
	\end{center}
	\caption{
		\label{fig:fig5}
		The rigid body diagram of the quadrotor \cite{cite:b18}.
	}
\end{figure}

When the Roller-Quadrotor's four-spoke wheel makes contact with a flat surface and the quadrotor is configured horizontally, the vehicle enters the flight mode. During flight, the dynamics of the Roller-Quadrotor resemble those of a conventional quadrotor, ensuring familiar characteristics and behaviors.

In the following section, we will present the fundamental dynamic model of the quadrotor. During the vehicle's flight mode, propulsion is provided by four rotors. Let us denote the thrust and torque generated by the $i$-th rotor (where $i\in \{1, 2, 3, 4\}$) as $F_i$ and $\tau_i$, respectively. In the body-fixed frame $\ss$, we define the four control inputs $U_i$ incorporating $F_i$ and $\tau_i$ as described in \cite{cite:b18}:
\begin{equation}
	\begin{cases}
		\label{eq:eq4}
		U_{1} = F_{1} + F_{2} + F_{3} + F_{4}\\
		U_{2} = (F_{2} - F_{4})D/2\\
		U_{3} = (F_{3} - F_{1})D/2\\
		U_{4} = \tau_2 + \tau_4 - \tau_1 - \tau_3\\
	\end{cases}
\end{equation}
In the aforementioned equations, $D_q$ represents the diagonal propeller distance of the quadrotor.

Consequently, the fundamental dynamic model \cite{cite:b19} of the vehicle in flight mode can be formulated as follows:
\begin{equation}
	\label{eq:eq5}
	\begin{cases}
		m \ddot{\mathbf{r}}=\mathbf{R}_{\mathcal{I}}^{\mathcal{B}}\left(U_1 \mathbf{Z}_B\right)-m g \mathbf{Z}_I \\
		\mathbf{I} \ddot{\mathbf{q}}=\left[U_2, U_3, U_4\right]^T-S(\mathbf{G q}) \mathbf{I}(\mathbf{G q})
	\end{cases}
\end{equation}
In the above equation, $\mathbf{r}=[x, y, z]^T$ represents the position of the center of mass in the inertial coordinates $I$, while $\mathbf{q}=[\phi, \theta, \psi]^T$ denotes the attitude of the quadrotor. The term $G$ signifies the affine transformation from the attitude angles to the angular velocities. The parameters $m$ and $I$ correspond to the mass and moments of inertia of the quadrotor, respectively. The variable $g$ denotes the gravitational constant. Moreover, $\mathbf{R}_{\mathcal{I}}^{\mathcal{B}}$ denotes the transformation matrix between the inertial frame $I$ and the body-fixed frame $\mathcal{B}$ as illustrated in Fig. \ref{fig:fig5}. Additionally, $S(\cdot)$ represents the skew matrix representation of the corresponding vector.

In summary, with Eq. \ref{eq:eq1} and Eq. \ref{eq:eq2}, by inputting the direction and dshot $U_d$ values of the four rotors, we can obtain $U_i, i\in {1, 2, 3, 4}$ in Eq. \ref{eq:eq4}, and then use Eq. \ref{eq:eq5} to model the flight mode.

To summarize, utilizing Eq. \ref{eq:eq1} and Eq. \ref{eq:eq2}, we can obtain $U_i$ (where $i\in \{1, 2, 3, 4\}$) by providing the direction and dshot $U_d$ values for the four rotors, as expressed in Eq. \ref{eq:eq4}. Subsequently, Eq. \ref{eq:eq5} is employed to model the flight mode.

\subsection{Transition Model}
In this section, we will delve into the transition phase, a crucial characteristic of the Roller-Quadrotor. The transition involves transforming the quadrotor into a rolling platform, wherein the rotors aid in maneuvering rather than providing lift. This mode enables the Roller-Quadrotor to navigate flat surfaces with enhanced efficiency and traverse challenging terrains.

\begin{figure}[h]
	\begin{center}
		\includegraphics[width=1.0\columnwidth]{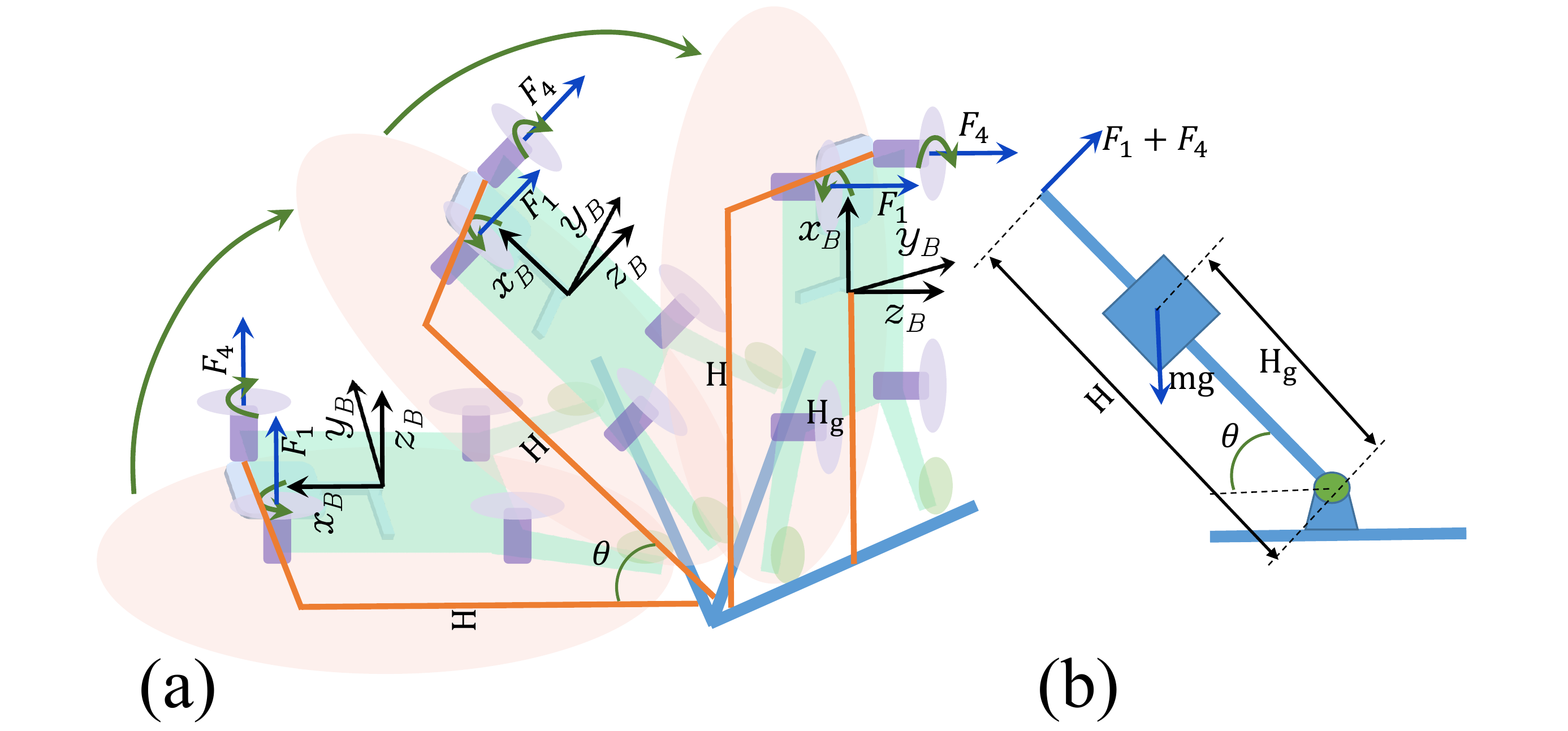}
	\end{center}
	\caption{
		\label{fig:fig6}
		Dynamic model of transition.
	}
\end{figure}

As illustrated in Fig. \ref{fig:fig6}, during the transition phase, the quadrotor is propelled by two rotors. By leveraging the optimization of the first-order inverted pendulum model \cite{cite:b16}, we develop a control model and controller for seamless transitions between aerial and terrestrial locomotion (refer to Fig. \ref{fig:fig6}(b)).
\begin{equation}
	\label{eq:eq6}
	\begin{cases}
		m l^2 \ddot{\theta}+m l g \cos (\theta)=\tau\\
		l = H_g\\
		\tau = (F_1 + F_4)H
	\end{cases}
\end{equation}
\begin{equation}
	\label{eq:eq7}
	\tau=\alpha \tau^{\prime}+\beta
\end{equation}
When we define $\alpha=m l^2$ and $\beta=m l g \cos(\theta)$, the system can be equivalently represented as $\tau^{\prime}=\ddot{\theta}$. The output of the controller is given by:
\begin{equation}
	\label{eq:eq8}
	\tau_1^{\prime}=\ddot{\theta}_d+k_v\left(\dot{\theta}_d-\dot{\theta}\right)+k_p\left(\theta_d-\theta\right)
\end{equation}
The control block diagram is depicted as follows:
\begin{figure}[h]
	\begin{center}
		\includegraphics[width=1.0\columnwidth]{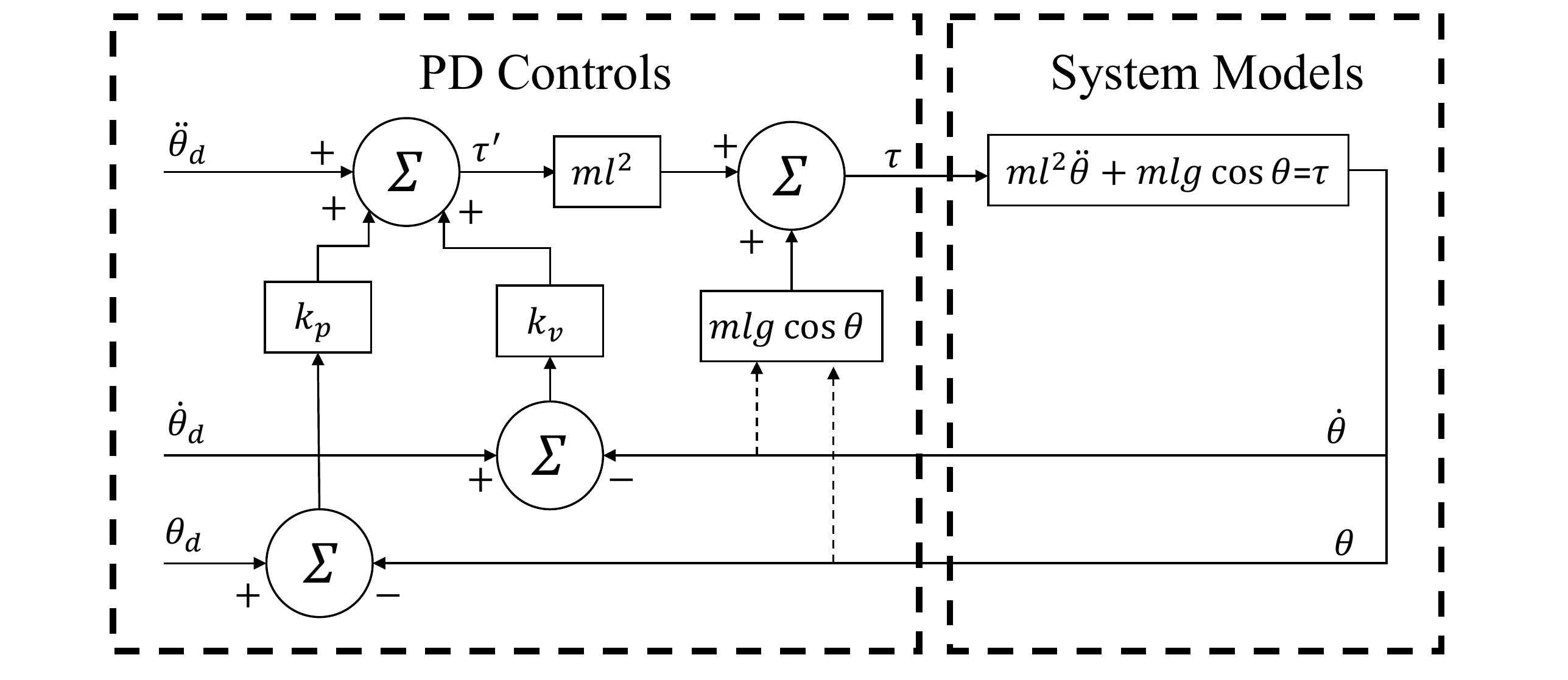}
	\end{center}
	\caption{
		\label{fig:fig7}
		Control block diagram of transition.
	}
\end{figure}

\subsection{Terrestrial Locomotion Model}
In this section, our attention will be directed towards the terrestrial locomotion of the Roller-Quadrotor. This locomotion mode employs a rolling motion to facilitate forward and backward movement, as well as turning. The utilization of this mode proves particularly advantageous when navigating through restricted spaces, overcoming obstacles, and executing tasks that demand accurate positioning and manipulation.

\begin{figure}[h]
	\begin{center}
		\includegraphics[width=1.0\columnwidth]{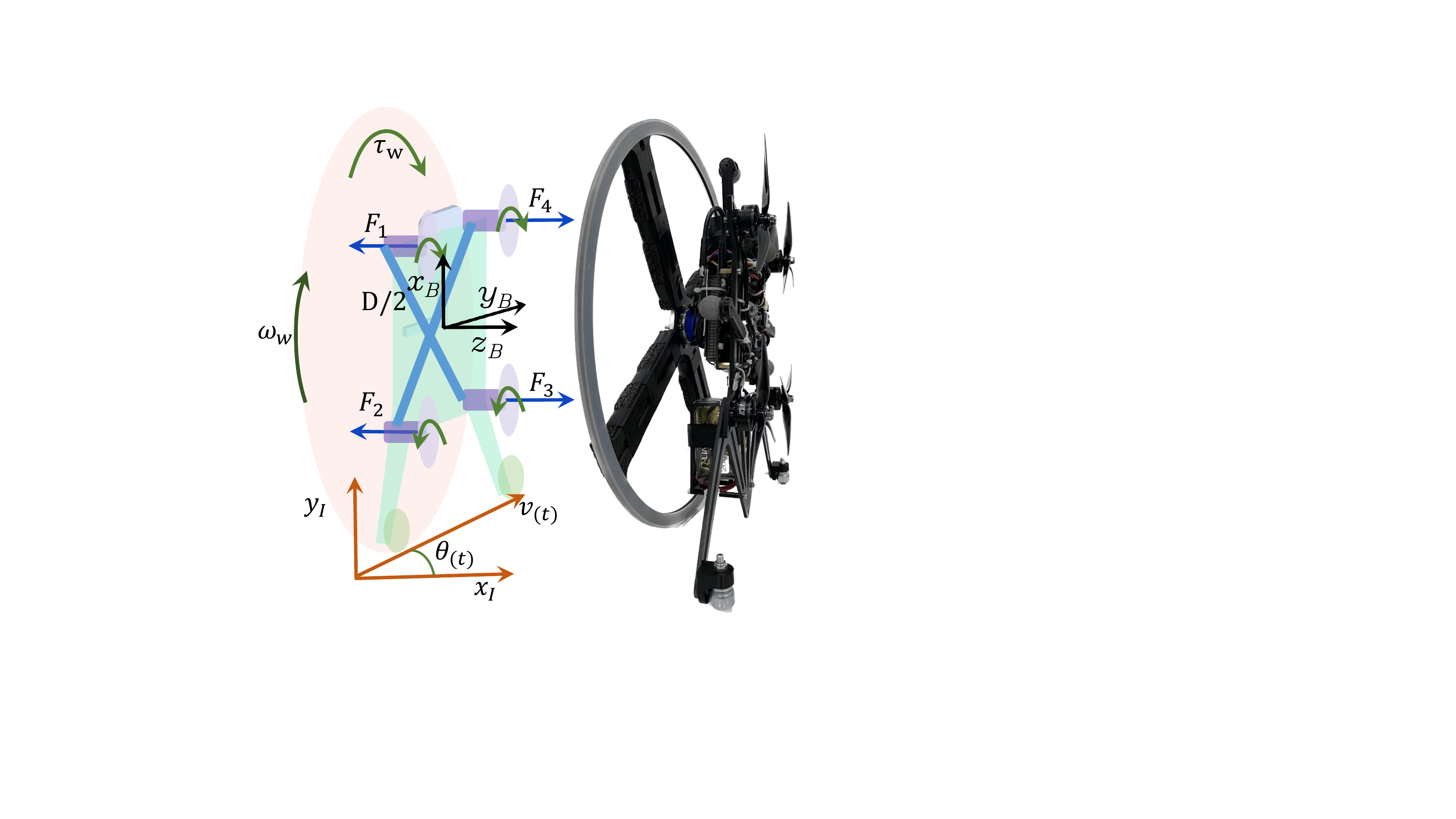}
	\end{center}
	\caption{
		\label{fig:fig8}
		Dynamic model of terrestrial locomotion.
	}
\end{figure}

We have developed a model and controller for the rolling motion of the vehicle, leveraging the optimization of the motion model of the planar unicycle model \cite{cite:b17}. As depicted in Fig. \ref{fig:fig8}, during the rolling mode, the quadrotor is propelled by one servomotor and four rotors.

Let's define the thrust generated by the $i$-th rotor (where $i\in\{1, 2, 3, 4\}$) as $F_i$. Additionally, we have the control over the angular velocity ($\omega_{w}$), linear velocity ($\nu_{w}$), torque ($\tau_{w}$), and force ($F_{w}$) of the wheel. We can directly manipulate and regulate these parameters using Eq. \ref{eq:eq1}, Eq. \ref{eq:eq2}, and Eq. \ref{eq:eq3}.

Let's define the position of the vehicle at time $t$ as $p_{x(t)}, p_{y(t)}, p_{z(t)}$, where $p_{z(t)} = 0$ since we are assuming a flat surface. Additionally, we define the linear velocity and yaw of the vehicle at time $t$ as $\nu_{(t)}$ and $\theta_{(t)}$ respectively.
\begin{equation}
	\label{eq:eq9}
	\tau =I \beta=I \frac{d \omega}{d t} = (F_1 + F_2-F_3 - F_4)\sqrt{2}D/4
\end{equation}
Based on the defined variables, we can model the state-space as follows:
\begin{equation}
	\label{eq:eq10}
	x_{1(t)} = p_{x(t)}, x_{2(t)} = p_{y(t)}, x_{3(t)} = \nu_{(t)}, x_{4(t)} = \theta_{(t)}\\
\end{equation}
\begin{equation}
	\label{eq:eq11}
	\vec{x}_{(t)}=\left[\begin{array}{c}p_{x(t)}, p_{y(t)}, \nu_{(t)}, \theta_{(t)}\end{array}\right]
\end{equation}
In the body-fixed frame $\ss$, we can define two control inputs $U_i$ as follows:
\begin{equation}
	\label{eq:eq12}
	U_{1} = \alpha_{(t)}, U_{2} = \omega_{(t)}
\end{equation}
\begin{equation}
	\label{eq:eq13}
	\vec{u}(t)=\left[\begin{array}{l}
		u_1(t) \\
		u_2(t)
	\end{array}\right]=\left[\begin{array}{l}
		\alpha(t) \\
		\omega_{(t)}
	\end{array}\right]
\end{equation}
Where $\alpha(t)$ represents the linear acceleration and $\omega_{(t)}$ represents the angular velocity of yaw.	
We establish the state-space equation as follows:
\begin{equation}
	\label{eq:eq14}
	\frac{d \vec{x}(t)}{d t}=\left[\begin{array}{c}
		\nu_{(t)} \cos \theta(t) \\
		\nu_{(t)} \sin \theta(t) \\
		0 \\
		0
	\end{array}\right]+\left[\begin{array}{c}
		0 \\
		0 \\
		\alpha(t) \\
		\omega(t)
	\end{array}\right]=f(\vec{x}(t), \vec{u}(t))
\end{equation}
When providing the desired trajectory, we can define it as follows:
\begin{equation}
	\label{eq:eq15}
	\left[\begin{array}{l}
		x_{1d}[k] \\
		x_{2d}[k]
	\end{array}\right]=\left[\begin{array}{l}
		p_{xd}[k] \\
		p_{yd}[k]
	\end{array}\right]
\end{equation}
\begin{equation}
	\label{eq:eq16}
	J=\left\|\vec{x}[N]-\vec{x}_d[N]\right\|_S^2+\sum_{k=1}^{N-1}\left(\left\|\vec{x}_{[k]}-\vec{x}_d[k]\right\|_Q^2+\left\|\vec{u}_{(k)}\right\|_R^2\right)
\end{equation}
where $J$ represents the performance measure, and $S$, $Q$, and $R$ are weight coefficient matrices. The goal is to optimize the control inputs $U^*$ in order to minimize the value of $J$.

\section{EXPERIMENTAL VALIDATION}




\begin{figure}[h]
	\begin{center}
		\includegraphics[width=1.0\columnwidth]{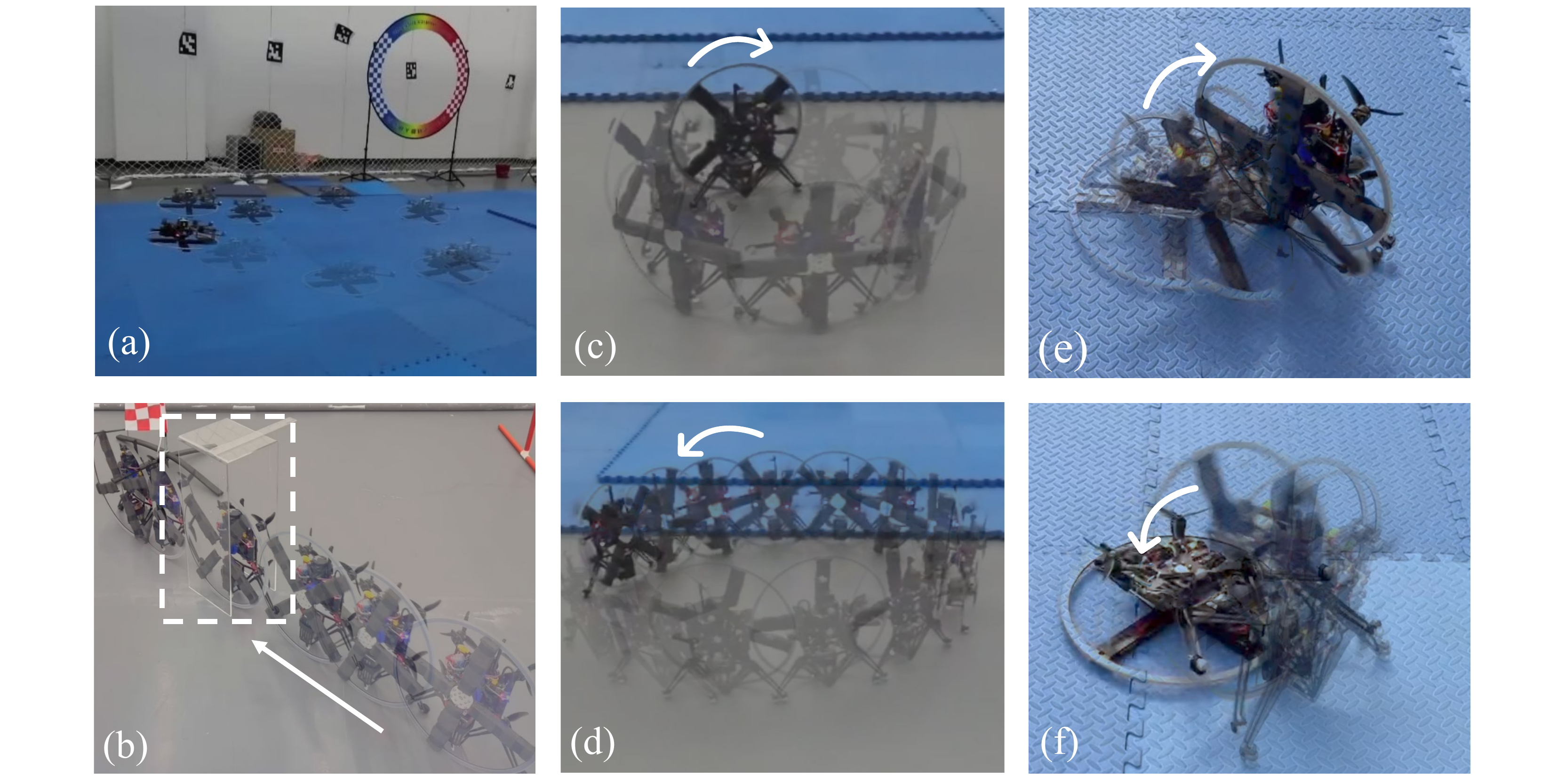}
	\end{center}
	\caption{
		\label{fig:fig9}
		Residual image: (a) the hand-flying experiment.
		(b) the experiment of maneuvering through a narrow 18 $cm$ gap.
		(c) the experiment of clockwise rolling circle motion.
		(d) the experiment of counterclockwise rolling circle motion.
		(e) the transition from flying to rolling mode.
		(f) the transition from rolling to flying mode.
	}
\end{figure}

\subsection{Experimental Setup}
In our real-world experiments, we conducted the tests within a controlled environment using a motion capture gym measuring $18\,m * 9\,m * 5\,m$.  This facility is equipped with a total of 26 Vicon cameras, which accurately capture the position and orientation of the quadrotor.  The state estimation of the quadrotor is obtained through an Extended Kalman Filter (EKF) that combines the pose information obtained from the Vicon cameras with the inertial measurement unit (IMU) data provided by the APM autopilot.

\subsection{Experiments of Taking Off and Flying}
Experiment 1: Hand-flying Experiments in Stabilize Mode.
To evaluate the flight performance of the Roller-Quadrotor, we conducted a series of hand-flying experiments in the Stabilize mode of the APM flight controller. The objective of this experiment was to assess the vehicle's ability to track the desired roll and pitch angles accurately over a 30-second flight duration.

\begin{figure}[h]
	\begin{center}
		\includegraphics[width=1.0\columnwidth]{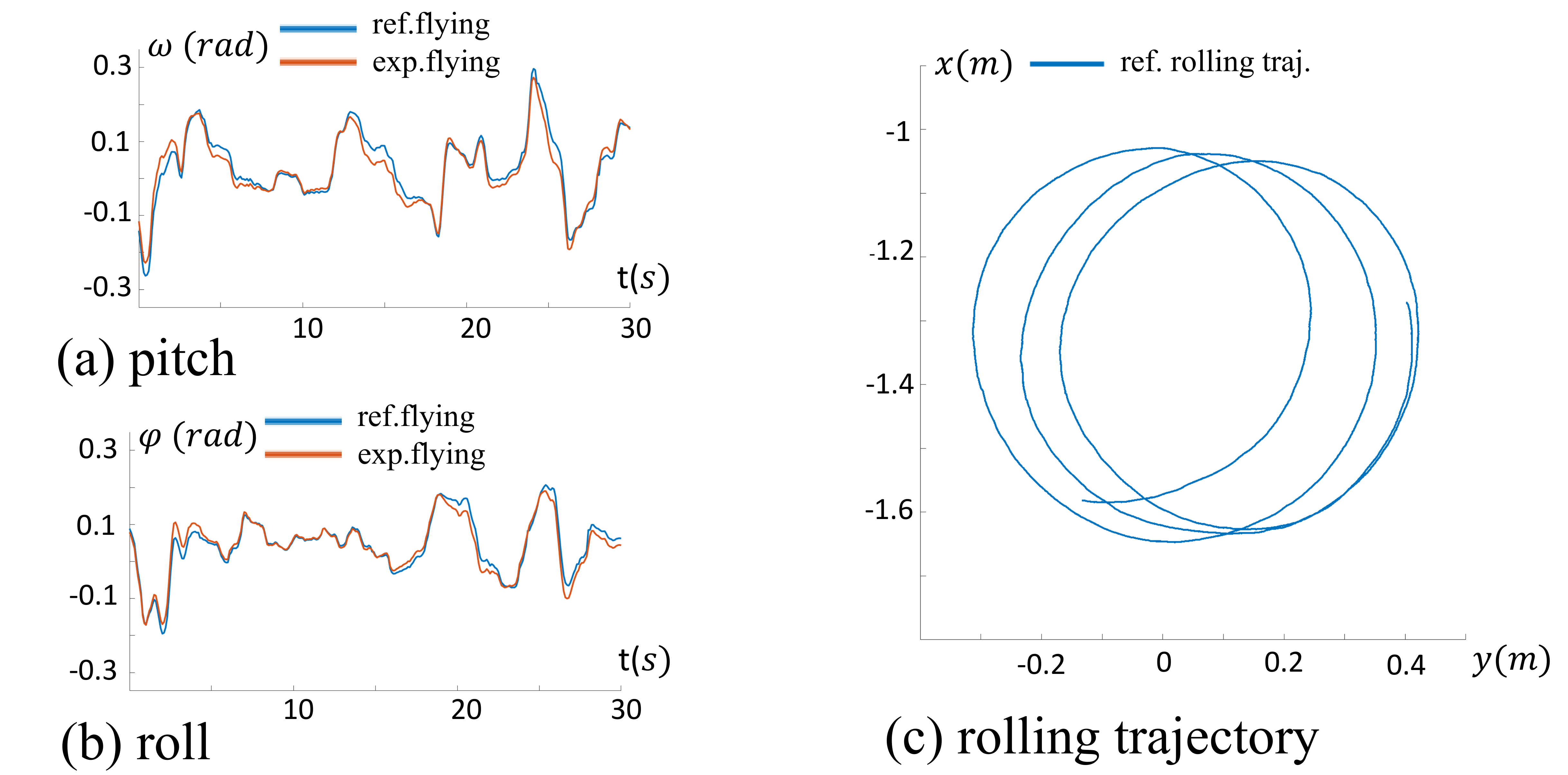}
	\end{center}
	\caption{
		\label{fig:fig10}
		(a) and (b) depict the roll and pitch angle tracking of the vehicle during manual flight in the stabilize mode, employing the APM flight controller. (c) presents a plot illustrating the vehicle's rolling trajectory during ground circle rolling.
	}
\end{figure}

During the experiment, we set the expected roll angle to be within 0.3 $rad$, and the expected pitch angle also within 0.3 $rad$. The maximum tracking error observed for the roll angle was 0.0555 $rad$, as depicted in Fig. \ref{fig:fig9} (a). Similarly, the maximum tracking error for the pitch angle was 0.0440 $rad$, as illustrated in Fig. \ref{fig:fig10} (a).

The results of Experiment 1 demonstrate the capability of the Roller-Quadrotor to achieve stable flight and accurately track the desired roll and pitch angles. These findings indicate that the vehicle exhibits satisfactory flight performance, paving the way for further experiments and validating its suitability for practical applications.

\subsection{Experiments of Transition}
Experiment 2 and 3: Modal Transition Experiments
In order to investigate the modal transition capabilities of the Roller-Quadrotor, we conducted experiments involving the transition from flying to rolling and from rolling to flying. These experiments were carried out on a foam floor, as depicted in Fig. \ref{fig:fig9} (e) and (f).

\subsection{Experiments of Terrestrial Locomotion}

\subsubsection{Pass Through Special Terrain}
Experiment 4: Narrow Gap Passage. We designed and constructed a gap/tunnel using acrylic material, with dimensions of $20\, cm * 18\, cm * 50\, cm$. The opening of the gap/tunnel was set to 18 $cm$ (as indicated by the white box in Fig. \ref{fig:fig9} (b)). The objective was to assess the Roller-Quadrotor's ability to navigate through narrow gaps in its rolling mode.

Through meticulous experimentation, we achieved success in maneuvering the vehicle through the narrow gap of 18 $cm$, given its 36 $cm$ diameter. This accomplishment demonstrates the Roller-Quadrotor's exceptional capability to navigate through challenging environments with restricted spaces. Notably, in a previous work by H. Jia et al. \cite{cite:b20}, a quadrotor with a passively reconfigurable airframe for hybrid terrestrial locomotion was able to pass through a narrow gap of 10 $cm$ with an 18 $cm$ diameter. In comparison, our Roller-Quadrotor, with the same diameter, effectively traversed a narrower gap of 9 $cm$ in a diameter of 18 $cm$. This improved performance showcases the superior maneuverability and agility of our vehicle.

The successful passage of the narrow gap validates the Roller-Quadrotor's enhanced terrain adaptability, specifically its ability to navigate through confined spaces. These results demonstrate the potential of our vehicle in various practical applications that require agile locomotion through narrow gaps and challenging environments.

\subsubsection{Ground Circle Rolling}
Experiment 5.1 and 5.2: Circular Track Roll

In Experiment 5.1, we set the experimental conditions with a circular track radius of 25 $cm$ in a clockwise direction (as depicted in Fig. \ref{fig:fig9} (c)). Similarly, in Experiment 5.2, the circular track radius was set to 50 $cm$ in a counterclockwise direction (as shown in Fig. \ref{fig:fig9} (d)). In both experiments, we programmed the expected yaw angle and allowed the vehicle to autonomously track the desired trajectory.

In Experiment 5.1, as illustrated in Fig. \ref{fig:fig10} (c), the vehicle successfully rolled for approximately 40 seconds. However, it should be noted that the trajectory exhibited some irregularities and lacked smoothness. Similar observations were made in Experiment 5.2.

These results indicate that while the Roller-Quadrotor demonstrated the ability to perform circular rolling motions, there is room for improvement in terms of trajectory smoothness. Further refinements and optimizations are necessary to enhance the vehicle's performance in executing precise and smooth circular rolling maneuvers.


\subsection{Experiments of Energy Consumption}
Experiment 6: Energy Consumption Analysis. We focused on evaluating the energy consumption of Roller-Quadrotor. Specifically, we set the vehicle to operate in the rolling mode with a fixed angular velocity of $\omega_{w} = 1\, rad/s$ and a circular trajectory radius of 25 $cm$ in a clockwise direction. The vehicle was allowed to run for an extended period of time under these conditions, during which we collected energy consumption data.

In the field of hybrid terrestrial and aerial quadrotors, Kalantari introduced a quadrotor named HyTAQ \cite{cite:b7} \cite{cite:b8}, which incorporates a rolling cage to enable terrestrial locomotion. To compare the energy-saving capabilities of Roller-Quadrotor, we present the results of the energy consumption experiments in Table \ref{tab:tab1}. These results demonstrate the remarkable energy efficiency of Roller-Quadrotor. It is worth noting that Roller-Quadrotor has a weight of approximately 1.5 $kg$, while HyTAQ weighs around 0.45 $kg$.

Analyzing the rolling energy consumption, Roller-Quadrotor exhibits a terrestrial range approximately 2.8 times greater and an operating time about 41.2 times greater than its aerial range/operating time. Furthermore, when compared to HyTAQ, Roller-Quadrotor achieves an operating time approximately 3.7 times greater in rolling mode, despite covering a shorter rolling distance. These comparisons are made under the assumption of equal vehicle mass.

These findings demonstrate the superior energy-saving performance of Roller-Quadrotor, emphasizing its suitability for extended terrestrial operations while preserving a favorable operational duration. The significant improvement in energy efficiency compared to existing quadrotor designs highlights the potential benefits and advantages of integrating rolling locomotion capabilities into aerial vehicles.
\begin{table}[ht]
	\renewcommand\arraystretch{1.2}
	\setlength{\tabcolsep}{0.15cm}
	\centering
	\caption{Results of Energy Consumption Experiments}
	\label{tab:tab1}
	\begin{tabular}{ccccc}
		\toprule
		\multicolumn{2}{c}{} & \makecell{Experiment Situation \\Time \\ Distance} & \makecell{Total\\ Energy \\ Consumption} & 		\makecell{Average Energy \\ Consumption\\Per Unit $kg$} \\
		\midrule
		\multicolumn{2}{c}{\makecell{Roller-Quadrotor\\Group 1} } &\makecell{Rolling \\ 48 $min$ \\ 518.4 $m$} & \makecell{1270 $mah$ \\ 14.8 $V$} & \makecell{15.7 $W$ \\87.0 $J/m$}\\
		\multicolumn{2}{c}{\makecell{Roller-Quadrotor\\Group 2} } & \makecell{Rolling \\ 40.33 $min$ \\ 435.6 $m$} & \makecell{1060 $mah$ \\ 14.8 $V$} & \makecell{15.6 $W$ \\86.4 $J/m$}\\
		\multicolumn{2}{c}{\makecell{Average of \\ Group 1 and 2} } &/ & / & \makecell{15.6 $W$ \\ 86.7 $J/m$} \\
		\multicolumn{2}{c}{\makecell{Roller-Quadrotor\\Group 3} } & \makecell{Manual flying \\ about 1.8 $min$ \\ 216 $m$} & \makecell{2000 $mah$ \\ 14.8 $V$} & \makecell{657.8 $W$ \\328.9 $J/m$}\\
		\multicolumn{2}{c}{\makecell{HyTAQ \cite{cite:b7} \cite{cite:b8}} } &\makecell{Rolling \\ 27 $min$ \\ 2400 $m$} & \makecell{1350 $mah$ \\ 11.1 $V$}& \makecell{74.0 $W$ \\50.0 $J/m$}\\
		\bottomrule
	\end{tabular}
\end{table}	
\subsection{Experiments of Hybrid Trrestrial and Aerial}
Experiment 7: Comprehensive Experimental Analysis. We conducted a series of comprehensive experiments to evaluate the performance of Roller-Quadrotor in various operational modes. The experiments included rolling locomotion, the transition from rolling to flight, take-off, and aerial flight (refer to Fig. \ref{fig:fig1}). The experimental setup aimed to simulate scenarios where the vehicle encounters obstacles during wheeled ground movement. Roller-Quadrotor demonstrated its capability to seamlessly transition from rolling to flight mode, enabling it to overcome obstacles by flying over them. This successful transition addresses the obstacle avoidance challenges typically encountered by terrestrial robots.

The experimental procedure involved initiating the vehicle in rolling mode, where it performed circular rolling motion in a clockwise direction with a desired diameter of 50 cm. Subsequently, the vehicle smoothly transitioned from rolling mode to flight mode. Finally, it executed a controlled take-off and continued to maneuver in the aerial domain.

These comprehensive experiments highlight the versatility and adaptability of Roller-Quadrotor, showcasing its ability to switch between ground and aerial modes seamlessly. By combining rolling and flight capabilities, the vehicle effectively addresses the limitations posed by obstacles encountered during terrestrial locomotion. Roller-Quadrotor's multifunctionality and obstacle-avoidance capabilities make it a promising solution for various real-world applications where both ground and aerial operations are required.

\section{CONCLUSION}
The Roller-Quadrotor is an innovative hybrid aerial-ground vehicle that combines the agility of quadrotors with the endurance of ground vehicles. This research study focuses on the design, modeling, and experimental validation of the Roller-Quadrotor. Flight capabilities are achieved through a quadrotor configuration, employing four thrust-providing actuators. Furthermore, rolling motion is facilitated by a unicycle-driven structure, augmented by rotor assistance for turning. By leveraging terrestrial locomotion, the vehicle effectively mitigates rolling and turning resistance, leading to energy conservation compared to its flight mode. This pioneering approach not only addresses the inherent challenges of conventional rotorcraft but also harnesses the vehicle's aerial mobility to navigate narrow gaps and surmount obstacles. We develop comprehensive models and controllers for the Roller-Quadrotor and validate their performance through experimental evaluations. The results demonstrate the seamless transition between aerial and terrestrial locomotion, showcasing the vehicle's capability to navigate safely through gaps half the size of its diameter. Moreover, the terrestrial range of the vehicle is approximately 2.8 times greater, accompanied by an operating time approximately 41.2 times longer compared to its aerial capabilities. These findings underscore the feasibility and efficacy of the proposed structure and control mechanisms, facilitating efficient navigation through challenging terrains while conserving energy.

In future endeavors, our research will prioritize enhancing model accuracy and developing sophisticated control algorithms.  These advancements aim to improving trajectory tracking accuracy.  Additionally, we intend to explore structural optimization techniques and weight reduction strategies to further enhance energy consumption performance.

\section{Acknowledgment}
The authors would like to thank Prof. Hao Li and Li Xu for their valuable suggestions.

\vspace{12pt}

\end{document}